\documentstyle[aclap]{article}

\newcommand{\hz}{\tt}

\title{\vspace{-0.5in}An Iterative Algorithm to Build Chinese Language Models}
\author{Xiaoqiang Luo \\
Center for Language \\ and Speech Processing\\ The Johns Hopkins University \\ 
3400 N. Charles St. \\ Baltimore, MD21218, USA \\{\tt xiao@jhu.edu} \And
Salim Roukos \\
IBM T. J. Watson Research Center \\
Yorktown Heights, NY 10598, USA \\
{\tt roukos@watson.ibm.com}
}

\begin{document}
\bibliographystyle{fullname}
\maketitle
\vspace{-0.5in}

\begin{abstract}
We present an iterative procedure to build a Chinese language model
(LM).  We segment Chinese text into words based on a word-based
Chinese language model. However, the construction of a Chinese LM
itself requires word boundaries. To get out of the chicken-and-egg
problem, we propose an iterative procedure that alternates two
operations:  segmenting text into words and building an LM. Starting
with an initial segmented corpus and an LM based upon it, we use a
Viterbi-liek algorithm to segment another set of data. Then, we build an LM
based on the second set and use the resulting LM to segment again the
first corpus. The alternating procedure provides a self-organized way
for the segmenter to detect automatically unseen words and correct
segmentation errors. Our preliminary experiment shows that the
alternating procedure not only improves the accuracy of our
segmentation, but discovers unseen words surprisingly well. The
resulting word-based LM has a perplexity of 188 for a general Chinese
corpus.
\end{abstract}

\section{Introduction}
In statistical speech recognition\cite{Bahl+al:83}, it is
necessary to build a language model(LM)
for assigning probabilities to hypothesized sentences.
The LM is usually built by collecting statistics
of words over a large set of text data. While doing so
is straightforward for English, it is not trivial to collect
statistics for Chinese words
since word boundaries are not marked in written Chinese text. Chinese
is a morphosyllabic language \cite{DeFrancis:84} in that almost all
Chinese characters represent a single syllable and most Chinese
characters are also morphemes. Since a word can be multi-syllabic, it
is generally non-trivial to segment a Chinese sentence into
words\cite{Wu+Tseng:93}. Since segmentation is a fundamental problem
in Chinese information processing, there is a large literature to deal
with the problem. Recent work includes \cite{Sproat+al:94} and
\cite{Wang+al:92}. In this paper, we adopt a statistical approach to
segment Chinese text based on an LM because of its autonomous nature
and its capability to handle unseen words.

As far as speech recognition is concerned, what is needed
is a model to assign a probability to a string of characters.
One may argue that we could bypass the segmentation problem by
building a character-based LM. However, we have a strong belief
that a word-based LM would be better than a character-based\footnote{
A character-based trigram model has a perplexity of
$46$ per character or $46^2$ per word (a Chinese word
has an average length of 2 characters), while a word-based trigram
model has a perplexity $188$ on the same set of data. While the
comparison would be fairer using a 5-gram character model, 
that the word model would have a lower perplexity as long as the
coverage is high.} one. In addition to speech recognition, the use of
word based models would have value in information retrieval and other
language processing applications.

If word boundaries are given, all established techniques can be
exploited to construct  an LM \cite{Jelinek+al:92}
just as is done for English. Therefore, segmentation is  a key issue in
building the Chinese LM. In this paper, we propose a segmentation
algorithm based on an LM. Since building an LM itself needs word
boundaries, this is a chicken-and-egg problem. To get out of this, we
propose an iterative procedure that alternates between the
segmentation of Chinese text and the construction of the LM. Our
preliminary experiments show that the iterative procedure is able to
improve the segmentation accuracy and more importantly, it can detect
unseen words automatically.

In section~\ref{seg-algorithm}, the Viterbi-like segmentation algorithm
based on a LM is described. Then in section~{section:iter-proc} we discuss
the alternating procedure of segmentation and building Chinese LMs. We test 
the segmentation algorithm and the alternating procedure and the results are 
reported in section~\ref{section:result}. Finally, the work is summarized in 
section~\ref{section:summary}.


\section{segmentation based on LM}
\label{seg-algorithm}
In this section, we assume there is a word-based Chinese LM at our disposal
so that we are able to compute the probability of a sentence (with
word boundaries). We use a Viterbi-like segmentation algorithm based on
the LM to segment texts.

Denote a sentence $S$ by $C_{1}C_{2}\cdots C_{n-1}C_{n}$, where
each $C_{i}$ ($1 \leq i \leq n$ \} is a Chinese character. To segment a sentence
into words is to group these characters into words, i.e.
\begin{eqnarray}
     S &=& C_{1}C_{2}\cdots C_{n-1}C_{n} \\
       &=& (C_{1}\cdots C_{x_{1}})(C_{x_{1}+1}\cdots C_{x_{2}}) \\
       & & \cdots (C_{x_{m-1}+1}\cdots  C_{x_{m}}) \\
       &=& w_{1}w_{2}\cdots w_{m}
\end{eqnarray}
where $x_{k}$ is the index of the last character in $k^{th}$ word $w_{k}$,
i,e $w_{k}=C_{x_{k-1}+1}\cdots C_{x_{k}} ( k=1,2,\cdots,m)$, and
of course, $x_{0}=0, x_{m}=n$.

Note that a segmentation of the sentence $S$ can be uniquely represented
by an integer sequence $x_{1}, \cdots, x_{m}$, so we will denote  a
segmentation by its corresponding integer sequence thereafter. Let
\begin{eqnarray}
G(S)=\{(x_{1}\cdots x_{m}): 1\leq x_{1}  \leq \cdots \leq x_{m}, m\leq n \}
\end{eqnarray}
be the set of all possible segmentations of sentence $S$. Suppose a
word-based LM is given, then for a segmentation $g(S)=(x_{1}\cdots
x_{m}) \in G(S)$, we can assign a score to $g(S)$ by
\begin{eqnarray}
L(g(S)) &=& \log P_{g}(w_{1}\cdots w_{m}) \\
	&=& \sum_{i=1}^{m} \log P_{g}(w_{i}|h_{i})
\end{eqnarray}
where $w_{j}=C_{x_{j-1}+1}\cdots C_{x_{j}}(j=1,2,\cdots,m)$, and $h_{i}$ 
is understood as  the history words
$w_{1}\cdots w_{i-1}$. In this paper the trigram model\cite{Jelinek+al:92} is 
used and therefore $h_{i}=w_{i-2}w_{i-1}$ 

 Among all
possible segmentations, we pick the one  $g^{*}$ with the highest
score as our result. That is,
\begin{eqnarray}
g^{*} &=& arg\max_{g \in G(S)} L(g(S))  \\
      &=& arg\max_{g \in G(S)} \log P_{g}(w_{1}\cdots w_{m}) \label{optimal-seg}
\end{eqnarray}

Note the score depends on segmentation $g$ and this is emphasized by
the subscript in (\ref{optimal-seg}). The optimal segmentation $g^{*}$
can be obtained by dynamic programming.
 With a slight abuse of notation, let
$L(k)$ be the max accumulated score for the first $k$ characters.
$L(k)$ is defined for $k=1,2,\cdots,n$ with $L(1)=0$ and $L(g^{*})=L(n)$.
Given $\{L(i): 1\leq i \leq k-1 \} $, $L(k)$ can be computed 
recursively as follows:
\begin{eqnarray}
  L(k) = \max_{1\leq i \leq k-1}[L(i) + \log P(C_{i+1}\cdots C_{k} | h_{i})]
               \label{DP:recursion}
\end{eqnarray}
where $h_{i}$ is the history words ended with the $i^{th}$ character $C_{i}$.
At the end of the recursion, we need to trace back to find the segmentation points.
Therefore, it's necessary to record the segmentation points in (\ref{DP:recursion}).

Let $p(k)$ be the index of the last character in the preceding word. Then
\begin{eqnarray}
  p(k) = arg \max_{1\leq i \leq k-1}[L(i) + \log P(C_{i+1}\cdots C_{k} | h_{i})]
               \label{DP:index}
\end{eqnarray}
that is, $C_{p(k)+1}\cdots C_{k} $ comprises the last word of the optimal segmentation
up to the $k^{th}$ character.

  A typical example
of a six-character sentence is shown in table~\ref{table:example}.
Since $p(6)=4$, we know the last word in the optimal segmentation is
$C_{5}C_{6}$. Since $p(4)=3$, the second last word is $C_{4}$. So on
and so forth. The optimal segmentation for this sentence is
\mbox{ $(C_{1}) (C_{2}C_{3}) (C_{4}) (C_{5}C_{6})$ }.
\begin{table}[h]
\begin{center}
\caption{A segmentation example}
\label{table:example}
\begin{tabular}{c|cccccc} \hline\hline
chars	& $C_{1}$ & $C_{2}$  & $C_{3}$ & $C_{4}$ & $C_{5}$ & $C_{6}$  \\ \hline
k       &   1     & 2        &  3      &  4      & 5       & 6  \\
p(k)    &   0     & 1        &  1      &  3      & 3       & 4  \\ \hline
\end{tabular}
\end{center}
\end{table}

The searches in (\ref{DP:recursion}) and (\ref{DP:index}) are in general time-consuming.
 Since long words are very rare in Chinese(94\% words are with three or less
characters \cite{Wu+Tseng:93}), it won't hurt at all to limit the search space
in (\ref{DP:recursion}) and (\ref{DP:index}) by putting an upper bound(say, 10) to the
length of the exploring word, i.e, impose the constraint $i\geq max{1,k-d}$
in (\ref{DP:recursion}) and (\ref{DP:index}),
 where $d$ is the upper bound of Chinese word length.
This will speed the dynamic programming significantly for long sentences.

It is worth of pointing out that the algorithm in (\ref{DP:recursion}) and
 (\ref{DP:index})
could pick an unseen word(i.e, a word not included in the vocabulary on which the LM
is built on)
in the optimal segmentation provided LM assigns proper probabilities to
unseen words. This is the beauty of the algorithm that it is able to
handle unseen words automatically.

\section{Iterative procedure to build LM}
\label{section:iter-proc}

In the previous section, we assumed there exists a Chinese word LM at our
disposal. However, this is not true in reality. In this section,
we discuss an iterative procedure that builds LM and automatically appends
the unseen words to the current vocabulary.

The procedure first splits the data into two parts, set $T_{1}$ and $T_{2}$.
We start from an initial segmentation of the set $T_{1}$. This can be done,
for instance,
by a simple greedy algorithm described in \cite{Sproat+al:94}. With the segmented
$T_{1}$, we construct a $LM_{i}$ on it. Then we segment the set $T_{2}$ by using
the $LM_{i}$ and the algorithm described in section~\ref{seg-algorithm}.
At the same time, we keep a counter
for each unseen word in optimal segmentations and increment the counter whenever
its associated word appears in an optimal segmentation. This gives us a
measure to tell whether an unseen word is an accidental character string
or a real word not included in our vocabulary. The higher a counter is, the more
likely it is a word. After segmenting the set $T_{2}$,
we add to our vocabulary all unseen words with its counter greater than
a threshold $c$. Then we use the augmented vocabulary and construct another
$LM_{i+1}$ using the segmented $T_{2}$. The pattern is clear now:
$LM_{i+1}$ is used to segment the set $T_{1}$ again and the vocabulary is
further augmented.

To be more precise, the procedure can be written in pseudo code as follows.
\begin{description}
   \item[Step 0:] Initially segment the set $T_{1}$. \\
		  Construct an LM $LM_{0}$ with an initial vocabulary $V_{0}$. \\
		  set i=1.
   \item[Step 1:] Let j=i mod 2; \\
		  For each  sentence $S$ in the set $T_{j}$, do
			\begin{description}
			  \item[1.1]segment it using $LM_{i-1}$.
			  \item[1.2]for each unseen word in the optimal segmentation,
			             increment its counter by the number of times it
			            appears in the optimal segmentation.
			\end{description}
   \item[Step 2:] Let $A$=the set of unseen words with counter greater than $c$. \\
		  set $V_{i} = V_{i-1} \cup A$. \\
	          Construct another $LM_{i}$ using the segmented set $T_{j}$
		  and the vocabulary $V_{i}$.

   \item[Step 3:] i=i+1 and goto step 1.
		
\end{description}

Unseen words, most of which are proper nouns,
pose a serious problem to Chinese text segmentation.
In \cite{Sproat+al:94} a class based model was proposed to identify
personal names. In \cite{Wang+al:92}, a title driven method was used
to identify personal names. The iterative procedure proposed here
provides a self-organized way to detect unseen words, including
proper nouns. The advantage is that it needs little human intervention.
The procedure provides a chance for us to correct segmenting errors.

\section{Experiments and Evaluation}
\label{section:result}

\subsection{Segmentation Accuracy}

Our first attempt is to see how accurate the segmentation algorithm
proposed in section \ref{seg-algorithm} is. To this end, we
split the whole data set \footnote{The corpus has about 5 million
characters and is coarsely pre-segmented.}
into two parts, half for building LMs and half reserved for testing.
The trigram model used in this experiment is
the standard deleted interpolation model described in \cite{Jelinek+al:92}
with a vocabulary of 20K words.

Since we lack an objective criterion to measure the accuracy of a segmentation
system, we ask three native speakers to segment manually
100 sentences picked randomly from the test set
and compare them with segmentations by machine.
The result is summed in table~\ref{seg-accuracy}, where ORG stands for
the original segmentation, P1, P2 and P3 for three human subjects, and TRI and UNI
stand for the segmentations generated by trigram LM and unigram LM respectively.
The number reported here is the arithmetic average of recall and precision, as
was used in \cite{Sproat+al:94}, i.e.,
$1/2(\frac{n_c}{n_1}+\frac{n_c}{n_2})$, where $n_c$ is the number of
common words in both segmentations, $n_1$ and $n_2$ are the number of
words in each of the segmentations.

\begin{table}[h]
\begin{center}
\caption{Segmentation Accuracy}
\begin{tabular}{|c|c|c|c|c|c|c|} \hline\hline
	& ORG  & P1   & P2   & P3   & TRI  & UNI  \\ \hline
ORG     &      &      &      &      & 94.2 & 91.2  \\ \hline
P1      & 85.9 &      &      &      & 85.3 & 87.4  \\ \hline
P2      & 79.1 & 90.9 &      &      & 80.1 & 82.2  \\ \hline
P3      & 87.4 & 85.7 & 82.2 &      & 85.6 & 85.7  \\ \hline \hline
\end{tabular}
\label{seg-accuracy}
\end{center}
\end{table}

%

We can make a few remarks about the result in table~\ref{seg-accuracy}.
First of all, it is interesting to note that the agreement of
segmentations among human subjects is roughly at the same level of
that between human subjects and machine. This confirms what reported
in \cite{Sproat+al:94}. The major disagreement for human subjects
comes from compound words, phrases and suffices. Since we don't give
any specific instructions to human subjects, one of them tends to
group consistently phrases as words because he was implicitly using
semantics as his segmentation criterion. For example, he segments the
sentence \footnote{Here we use Pin Yin followed by its tone to
represent a character.}  {\hz dao4 jia1 li2 chi1 dun4 fan4}(see
table~\ref{tab:compound}) as two words {\hz dao4 jia1 li2(go home)}
and {\hz chi1 dun4 fan4(have a meal)} because the two ``words'' are
clearly two semantic units. The other two subjects and machine segment
it as {\hz dao4 / jia1 li2/ chi1/ dun4 / fan4}.

Chinese has very limited morphology \cite{Spencer:91} in that most
grammatical concepts are conveyed by separate words and not by
morphological processes. The limited morphology includes some ending
morphemes to represent tenses of verbs, and this is another source of
disagreement. For example, for the partial sentence {\hz zuo4 wan2
le}, where {\hz le} functions as labeling the verb {\hz zuo4 wan2} as
``perfect'' tense, some subjects tend to segment it as two words {\hz
zuo4 wan2/ le} while the other treat it as one single word.

Second, the agreement of each of the subjects with either the
original, trigram, or unigram segmentation is quite high (see columns
2, 6, and 7 in Table~\ref{seg-accuracy}) and appears to be specific to the subject.

\begin{table}[h]
\begin{center}
\caption{Segmentation of phrases}
\begin{tabular}{c|ccccc} \hline\hline
Chinese	&dao4 & jia1  li2  & chi1 & dun4 &  fan4 \\
Meaning &go   & home       &  eat & a  & meal \\ \hline \hline
\end{tabular}
\label{tab:compound}
\end{center}
\end{table}

Third, it seems puzzling that the trigram LM agrees with the original
segmentation better than a unigram model, but gives a worse result
when compared with manual segmentations. However,
since the LMs are trained using the presegmented data,
the trigram model tends to keep the original segmentation
because it takes the preceding two words into account while
the unigram model is less restricted to deviate from the original
segmentation. In other words, if trained with ``cleanly'' segmented
data, a trigram model is more likely to produce a better segmentation
since it tends to preserve the nature of training data.

\subsection{Experiment of the iterative procedure}
In addition to the 5 million characters of segmented text, we had
unsegmented data from various sources
 reaching about 13 million 
characters. We applied our iterative algorithm to that corpus.

\begin{table}[h]
\begin{center}
\caption{Segmentation of accuracy after one iteration} 
\begin{tabular}{|c|cc|} \hline\hline
	& TR0  & TR1 \\ 
ORG     & .920 & .890 \\
P1      & .863 & .877 \\
P2      & .817 & .832 \\
P3      & .850 & .849 \\ \hline \hline
\end{tabular}
\label{tab:accuracy-iter}
\end{center}
\end{table}
 Table~\ref{tab:accuracy-iter}
shows the figure of merit of the resulting segmentation of the 100
sentence test set described earlier. After one iteration, the agreement
with the original segmentation decreased by 3 percentage points, while
the  agreement with the human segmentation increased by less than one
percentage point. We ran our computation intensive procedure for
one iteration only. The results indicate that the impact on
segmentation accuracy would be small. However, the new unsegmented
corpus is a good source of automatically discovered words. A 20 
examples picked randomly from about 1500 unseen words 
are shown in Table~\ref{tab:unseen-words}. 
16 of them are 
reasonably good words and are listed with their 
translated meanings. The problematic words are marked with ``?''.

%
\begin{tiny}
\begin{table}[h]
\begin{center}
\caption{Examples of unseen words}\label{tab:unseen-words}
\begin{tabular}{r|l} \hline\hline
PinYin & Meaning  \\ \hline
kui2 er2 & last name of former US vice president  \\
he2 shi4 lu4 yin1 dai4 & 
				cassette of audio tape \\
shou2 dao3    & (abbr)pretect (the) island \\
ren4 zhong4   & first name or part of a phrase \\
ji4 jian3     & (abbr) discipline monitoring \\
zi4 hai4  &  ? \\
shuang1 bao3   & double guarantee  \\
ji4 dong1  & (abbr) Eastern He Bei province  \\
zi3 jiao1  & purple glue  \\
xiao1 long2 shi2  & personal name  \\
li4 bo4 hai3 & ?  \\
du4 shan1   & ?  \\
shang1 ban4 & (abbr) commercial oriented \\
liu6 hai4  & six (types of) harms  \\
sa4 he4 le4  & translated name  \\
kuai4 xun4 & fast news  \\
cheng4 jing3   & train cop  \\
huang2 du2  & yellow poison  \\
ba3 lian2  &  ? \\
he2 dao3  & a (biological) jargon \\
 \hline
\end{tabular}
\end{center}
\end{table}
\end{tiny}


\subsection{Perplexity of the language model}

After each segmentation, an interpolated trigram model is built, and 
an independent test set with 2.5 million characters is segmented
and then
used to measure the quality of the model. 
We got a perplexity 188 for a vocabulary of 80K words,
 and the alternating procedure has little impact 
on the perplexity. This can be explained by the fact that 
the change of segmentation 
is very little ( which is reflected in table~ref{tab:accuracy-iter} )
and the addition of unseen words(1.5K) to the vocabulary is also 
too little to affect the overall perplexity. The merit of 
the alternating procedure is probably its ability to detect unseen words. 

\section{Conclusion}
\label{section:summary}

In this paper, we present an iterative procedure to
build Chinese language model(LM).  We segment Chinese text into
words based on a word-based Chinese language model.
However, the construction of a Chinese LM itself requires
word boundaries.
To get out of the chicken-egg problem, we propose
an iterative procedure that alternates two operations:
segmenting text into words and
building an LM.
Starting with an
initial segmented corpus and an LM based upon it, we use Viterbi-like algorithm
to segment another set of data. Then we build
an LM based on the second set and use the LM to segment again
the first corpus. The alternating procedure
provides a self-organized way for the segmenter to detect automatically
unseen words and correct segmentation errors.
Our preliminary experiment shows that
the alternating procedure not only improves the accuracy of our segmentation,
but discovers unseen words surprisingly well. We get a perplexity 188
for a general Chinese corpus with 2.5 million characters
\footnote{Unfortunately, we could not find a report of Chinese 
perplexity for comparison in 
the published literature concerning Mandarin speech recognition}.

\section{Acknowledgment}
The first author would like to thank various members of the Human Language
technologies Department at the IBM T.J Watson center for their
encouragement and helpful advice.  Special thanks go to
Dr. Martin Franz for providing continuous help in using the IBM language
model tools. The authors would also thank the comments and insight of two anonymous 
reviewers which help improve the final draft.

\end{document}